# Response Prediction of Structural System Subject to Earthquake Motions using Artificial Neural Network

S. Chakraverty*, T. Marwala** , Pallavi Gupta* and Thando Tettey**

*B.P.P.P. Division, Central Building Research Institute
Roorkee-247 667, Uttaranchal, India
e-mail :sne_chak@yahoo.com

**School of Electrical and Information Engineering,
University of the Witwatersrand, Private Bag 3
Wits, 2050,Republic of South Africa

## Abstract

This paper uses Artificial Neural Network (ANN) models to compute response of structural system subject to Indian earthquakes at Chamoli and Uttarkashi ground motion data. The system is first trained for a single real earthquake data. The trained ANN architecture is then used to simulate earthquakes with various intensities and it was found that the predicted responses given by ANN model are accurate for practical purposes. When the ANN is trained by a part of the ground motion data, it can also identify the responses of the structural system well. In this way the safeness of the structural systems may be predicted in case of future earthquakes without waiting for the earthquake to occur for the lessons. Time period and the corresponding maximum response of the building for an earthquake has been evaluated, which is again trained to predict the maximum response of the building at different time periods. The trained time period versus maximum response ANN model is also tested for real earthquake data of other place, which was not used in the training and was found to be in good agreement.

**Keywords :** Earthquake, Neural Network, Frequency, Structure, Building.

# 1 Introduction

Real earthquake ground motion at a particular building site is very complicated. The response of a building to an earthquake is dynamic and for a dynamic response, the building is subjected to a vibratory shaking of the base. Exactly how a building responds is complex and depends on the amplitude and frequency of vibration along with the material and design of the building. All buildings have a "natural frequency" associated with them. If strain is placed on to the structure and then let it snap back into equilibrium, it will sway back and forth with an amplitude that decays with time. If the ground shakes with the same frequency as a building's natural frequency, it will cause the amplitude of sway to get larger and larger such that, the ground shaking is in resonance with the building's natural frequency. This produces the most strain on the components of the building and can quickly cause the building to collapse. Powerful technique of Artificial Neural Network (ANN) has been used to model the problem for one storey structure. Among the different types of ANN, the feedforward, multilayer, supervised neural network with error back propagation algorithm, the BPN [1] is the most frequently applied NN model. Dynamic response of a structure to strong earthquake ground motion may be investigated by different methods. The method, that has been used here, is to create a trained black box containing the characteristics of the structure and of the earthquake motion which can predict the dynamic response for any other earthquake for a particular structure.

Artificial Neural Network (ANN) have gradually been established as a powerful soft computing tool in various fields because of their excellent learning capacity and their high tolerance to partially inaccurate data. ANN has, recently been applied to assess damage in structures. Stefano et al.[3] used probabilistic Neural Networks for seismic damage prediction. Many methods viz. [4]-[9] were introduced for response estimation and for structural control. Zhao et al.[10]

applied a counter-propagation NN to locate damage in beams and frames. KuZniar and Waszczyszyn [11] simulated the dynamic response for prefabricated building using ANN. Elkordy et al.[12] used a back-propagation neural network with modal shapes in the input layer to detect the simulated damage of structures. Muhammad [13] gives certain ANN applications in concrete structures. Pandey and Barai [14] detected damage in a bridge truss by applying ANN of multilayer perceptron architectures to numerically simulated data. Some studies such as [15]-[17] used artificial neural network for structural damage detection and system identification.

In the present paper, the Chamoli earthquake ground acceleration at Barkot (NE) and Uttarkashi earthquake ground acceleration recorded at Barkot (NE and NW) have been considered based on the authors' previous study [18]. From their ground acceleration the responses are computed using the usual procedure. Then the ground acceleration and the corresponding response are trained using Artificial Neural Network (ANN) with and without damping. After training the network with one earthquake, the converged weight matrices are stored. In order to show the power of these converged (trained) networks other earthquakes are used as input to predict the direct response of the structure without using any mathematical analysis of the response prediction. Similarly, the various time periods of one earthquake and its corresponding maximum responses are trained. Then the converged weights are used to predict the maximum response directly to the corresponding time period. Various other results related to use of these trained networks are discussed for future / other earthquakes.

## 2      Artificial Neural Network

Artificial neural systems are present day machines that have great potential to improve the quality of our life. Advances have been made in applying such systems for problems found difficult for traditional computation. A neural network is a parallel, distributed information processing structure consists of processing elements called neurons, which are interconnected and unidirectional signal channels called connections. The general structure of the network that have been used here is given in Fig.1. the structure consists of three layers : the input layer, the hidden layer and the output layer. The input layer is made up of one or more neurons or processing elements that collectively represent the information in a particular pattern of a training set. The hidden layer also consists of one or more neurons. Its purpose is simply to transform the information from the input layer to prepare it for the output layer. The output layer, which has one or more neurons, uses input from the hidden layer (which is a transformation of the input layer) to produce an output value for the entire network. The output is used to interpret the training and classification results of the network. The processing elements or neurons are connected to each other by adjustable weights. The input/output behaviour of the network changes if the weights are changed. So, the weights of the net may be chosen in such a way so as to achieve a desired output. To satisfy this goal, systematic ways of adjusting the weights have to be developed, which are known as training or learning algorithm.

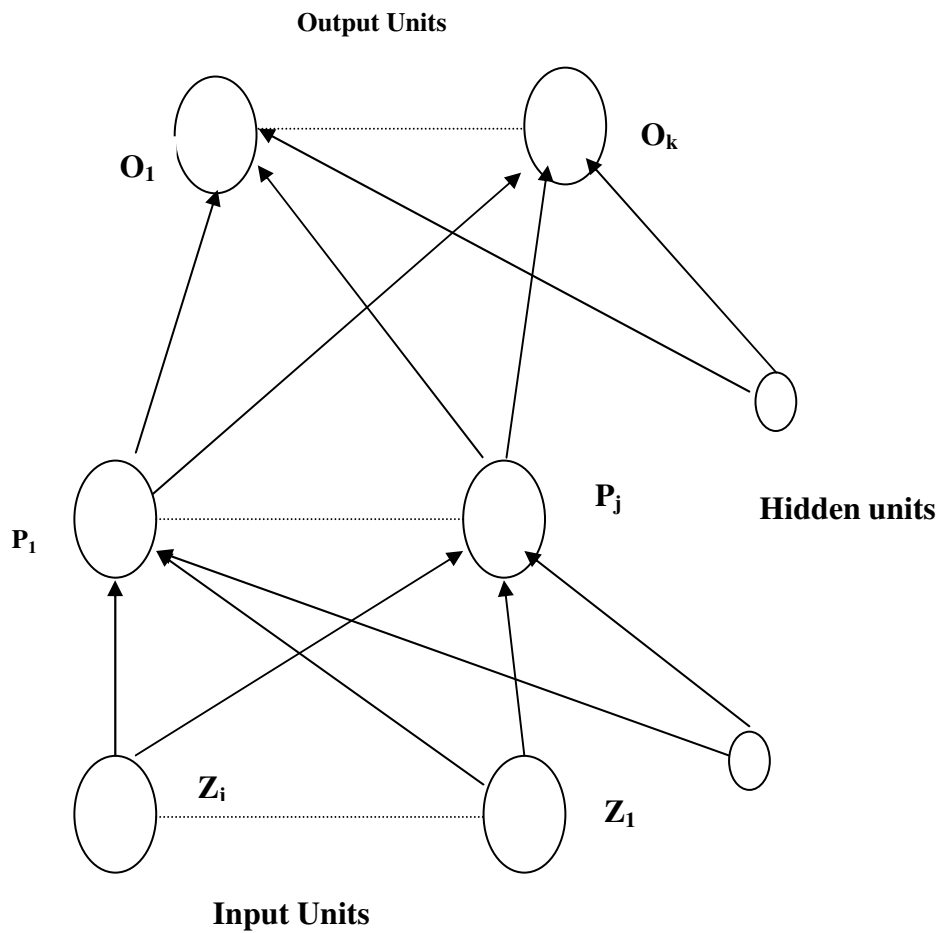

Output Units

$O_1$  $O_k$

$P_1$  $P_j$  Hidden units

$Z_i$  $Z_1$

Input Units

## 3     Error Back Propagation Training Algorithm (EBPTA)

Here, Error Back Propagation Training algorithm and feedforward recall with one hidden layer has been used. In Fig. 1, $Z_i$, $P_j$ and $O_k$ are input, hidden and output layer respectively. The weights between input and hidden layers are denoted by $v_{ji}$ and the weights between hidden and output layers are denoted by $W_{kj}$. The procedure may easily be written down for the processing of this algorithm.

Given R training pairs

$$\{Z_1, d_1; Z_2, d_2; \ldots\ldots Z_R, d_R\}$$

where $Z_i$ (Ix1) are input and $d_i$ (Kx1) are desired values for the given inputs, the error value is computed as

$$E = \frac{1}{2}(d_k - O_k)^2, \quad k = 1,2,\ldots K$$

for the present neural network as shown in Fig. 1.

The error signal terms of the output ($\delta_{Ok}$) and hidden layers ($\delta_{Pj}$) are written respectively as,

$$\delta_{Ok} = 0.5 * (d_k - O_k)(1 - O_k^2), \quad k = 1,2,\ldots\ldots K$$
$$\delta_{Pj} = 0.5 * (1 - P_j^2) \sum_{k=1}^{K} \delta_{Ok} W_{Pj}, \quad j = 1,2,\ldots J$$

Consequently, output layer weights ($W_{kj}$) and hidden layer weights ($\delta_{ji}$) are adjusted as,

$$v_{ji}^{(New)} = v_{ji}^{(Old)} + \beta \delta_{Pj} Z_i, \quad j = 1,2\ldots\ldots J \text{ and } i = 1,2,\ldots\ldots I$$

$$W_{kj}^{(New)} = W_{kj}^{(Old)} + \beta \delta_{Ok} P_j, \quad k = 1,2\ldots K \text{ and } j = 1,2,\ldots J$$

Where, β is the learning constant.

## 4   Response Prediction

The basic idea behind the proposed methodology is to predict the structural response of single degree of freedom system i.e. single storey building subject to various earthquake forces. Two cases viz without damping and with damping have been considered for the analysis.

**Case(i)** : Without damping

Let M be the mass of the generalized one storey structure, K the stiffness of the structure and x be the displacement relative to the ground then the equation of motion may be written as:

$$M\ddot{x} + Kx = -M\ddot{a} \qquad (1)$$

$\ddot{x}$ = Response acceleration,
x = Displacement,
$\ddot{a}$ = Ground acceleration.
where,

Equation (1) may be written as,

$$\ddot{x} + \omega^2 x = -\ddot{a} \qquad (2)$$

Where $\omega^2$=K/M, is the natural frequency parameter of the undamped structure.

The solution of equation (2) [Ref. 2] is given by

$$x(t) = -\frac{1}{\omega}\int_0^t \ddot{a}(\tau)\sin[\omega(t-\tau)]d\tau \qquad (3)$$

From this solution the response of the structure viz. acceleration is obtained for no damping.

**Case (ii)** : With damping

Let M be the mass of the generalized one storey structure, K the stiffness of the structure, C the damping and x be the displacement relative to the ground then the equation of motion may be written as:

$$M\ddot{x} + C\dot{x} + Kx = -M\ddot{a} \tag{4}$$

where

- $\ddot{x}$ = Response acceleration,
- $\dot{x}$ = Response velocity,
- $x$ = Displacement,
- $\ddot{a}$ = Ground acceleration.

Equation (4) may be written as,

$$\ddot{x} + 2\xi\omega\dot{x} + \omega^2 x = -\ddot{a} \tag{5}$$

Where $\xi\omega$ = C/2M and $\omega^2$=K/M, is the natural frequency parameter of the undamped structure.

The solution of equation (5) [Ref.2] is given by

$$x(t) = -\frac{1}{\omega}\int_0^t \ddot{a}(\tau)\exp[-\xi\omega(t-\tau)]\sin[\omega(t-\tau)]d\tau \tag{6}$$

From this solution the response of the structure viz. acceleration is obtained for damping.

Now, the neural network architecture is constructed, taking ground acceleration as input and the response obtained from the above solution is taken as output for each time step. Therefore, the whole network consists of one input layer, one hidden layer with varying nodes and one output layer as shown in Fig.1. Similarly for the other problem of time period vs. maximum response the input and output layer contain the time period and the corresponding maximum response respectively at each interval for the particular structure.

## 5 Numerical Results and Discussions

For the present study two Indian earthquakes viz. the Chamoli Earthquake (max. ground acceleration =0.16885 m/sec/sec) at Barkot in NE (north–east) direction shown in Fig.2(a) and the Uttarkashi earthquake (maximum ground acceleration = 0.931 m/sec/sec) at Barkot in NE (north–east) and NW (north-west) direction as given in Figs. 2(b) and 2(c) have been considered for training and testing.

Initially, the system without damping is studied and for that the ground acceleration of Chamoli earthquake at Barkot (NE) was used to compute the response for single storey structure using usual procedure from Eq.(3). The obtained response and the ground acceleration is trained first for the assumed frequency parameters $\omega=0.5$ and $\omega=0.01$ for time range 0 to 14.92 sec.(748 data points) for the mentioned earthquake. Simulations have been done for different hidden layer nodes and it was seen that the response result is almost same and good for 5 to 20 nodes in the hidden layer. However, 10 hidden layer nodes are used here to generate further results.

After training ground acceleration and response data for Chamoli earthquake at Barkot (NE) for 10 nodes in hidden layer, the weights are stored and they are used to predict responses for various intensity earthquakes. The plot in Fig. 3(a) shows response comparison between neural and desired for the 80% of Chamoli earthquake at Barkot (NE) for $\omega=0.01$(maximum response=0.135079m/sec/sec). Similarly, the response comparison for 120% Chamoli earthquake at Barkot (NE) for $\omega=0.5$ (Maximum response=0.20260 m/sec/sec) is shown in Fig 3(b).

Next, a part of the ground acceleration is used for the training and it will be shown that the present ANN can predict the whole period of the response using the trained ANN by the part of the data. So, the ground acceleration and response data with Chamoli earthquake is trained for an example with the time range 0 to 10.96 sec.(550 data points). Its weights are stored to find the response for the time range 0 to 14.92 sec. (whole period) at different percentages of the earthquake in order to test the network learning for the points outside the training set. Figs. 4(a) and 4(b) show the response comparison between neural and desired for $\omega$=0.01, (maximum response=0.168849 m/sec/sec) and for $\omega$=0.5 (maximum response=0.168841 m/sec/sec) at the time range 0 to 10.96 sec. The response comparison between neural and desired for $\omega$=0.01 with 120% of the earthquake force (maximum response = 0.20260 m/sec/sec) from the time range 0 to 14.92 sec.(748 data points) is incorporated in Fig. 5(a). It is obtained from the weights of the trained data for the time range 0 to 10.96 sec. (550 data points). From the same weights, neural responses for 80% of earthquake force, is computed with $\omega$=0.5(maximum response=0.135073 m/sec/sec), for the time range 0 to 14.92 sec. (748 data points) and it is plotted in Fig. 5(b).

The system with damping is then considered and for this, first from the ground acceleration of Chamoli Earthquake at Barkot (NE), the response is computed using Eq.(6). The obtained responses and the ground acceleration are trained by the said ANN model for an example structural system with frequency parameter $\omega$= 0.68981 and damping = 1.58033. This training was done for the total time range 0 to 14.92 sec. (748 points, earthquake period). Plot of 100% response comparison between neural and desired for Chamoli Earthquake at barkot (NE) is shown in Fig. 6(a). After training ground acceleration and response data for Chamoli Earthquake for various nodes in the hidden layer it was confirmed that 10 nodes are again sufficient for the prediction. So, the weights corresponding to 10 hidden nodes are stored and they are used to predict responses for various intensity earthquakes. The response for 50% ($\omega$= 0.68981, damping = 1.58033 and maximum response = 0.00375 m/sec/sec) of the Chamoli Earthquake at Barkot (NE) and its comparison with the desired response are shown in Fig. 6(b). Similarly, the response comparison between neural and desired is shown in Fig.

6(c) (ω= 0.68981, damping = 1.58033 and maximum response for 120% = 0.00910 m/sec/sec) for 120% of earthquake acceleration.

Finally, the Uttarkashi earthquake at Barkot (NW) ground acceleration is used with damping = 0.05, at different time periods ( t = 1/omega) ranging from 0.5 to 10 with an interval of 0.02 (620 data points) for evaluating the maximum responses corresponding to each time period using Eq. (6). The obtained time periods and the corresponding responses are trained and then the converged weights are stored. The comparison between neural and desired results is shown in Fig. 7(a). The stored weights were then used to predict the response for different time periods lying in the same range of 0.5 to 10 but at different time interval of 0.5 for another earthquake such as Uttarkashi earthquake at Tehri (NW), The results are depicted in  Fig. 7(b) showing good comparison between ANN model and desired results.

# 6     Conclusions

This paper uses the powerful soft computing technique (Artificial Neural Network) to compute structural response of single degree of freedom system subject to Indian earthquakes at Chamoli and Uttarkashi ground motion data. Also this technique is used to predict the maximum response corresponding to various time periods. It is shown here that once the training is done then the trained architecture may be used to simulate for various intensity earthquakes, thereby showing the responses of the system which depend upon the structural properties (mass and stiffness) of the structure. If the network is trained for various time periods of one earthquake and its corresponding maximum responses then the model can predict the maximum response directly to the corresponding time period for any other earthquake that had not been used during the training. In this way the safety of the structural systems may be predicted in case of future earthquakes.

**Acknowledgements**

The authors would like to thank Department of Science and Technology, India for funding and Director C.B.R.I. for giving permission to publish this paper.

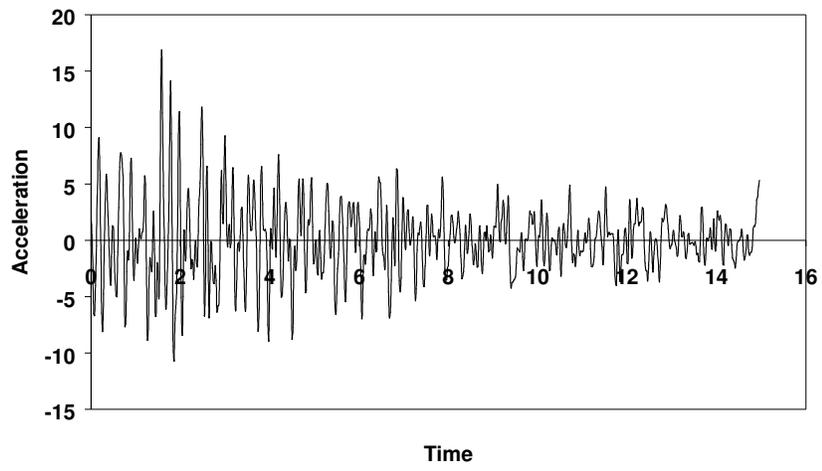

**Fig. 2(a). Chamoli Earthquake at Barkot in NE direction
Peak Acceleration = 0.16885m/sec/sec**

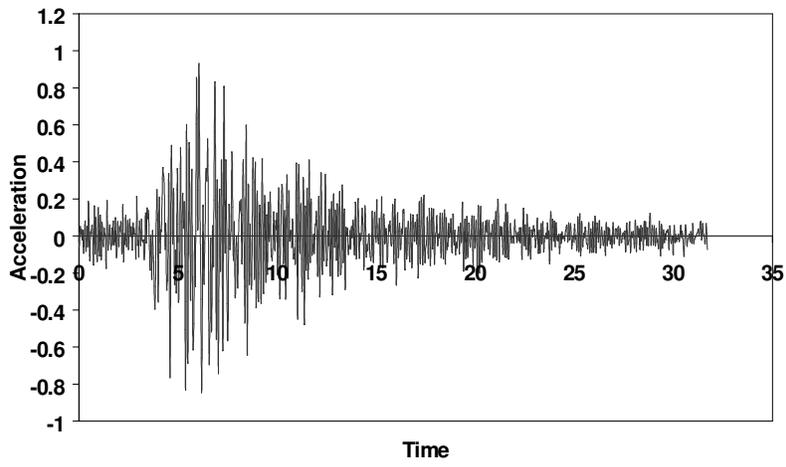

**Fig. 2(b) Uttarkashi Earthquake at Barkot in NE direction
Peak Acceleration : 0.9317m/sec/sec**

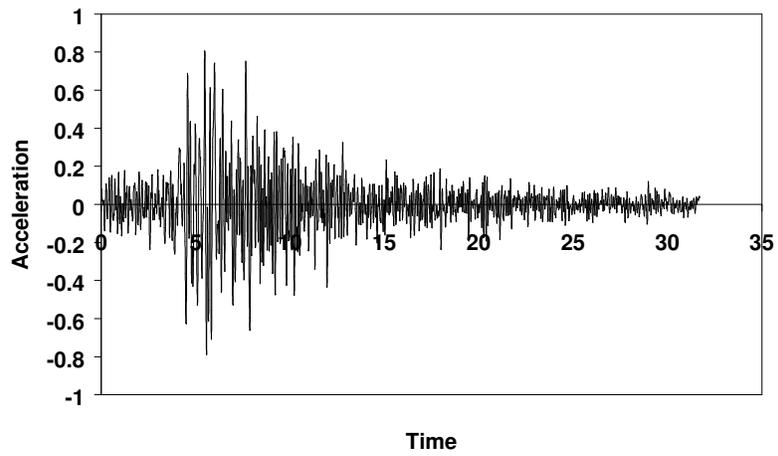

**Fig. 2(c). Uttarkashi Earthquake at Barkot in NW direction
Peak Acceleration : 0.8470m/sec/sec**

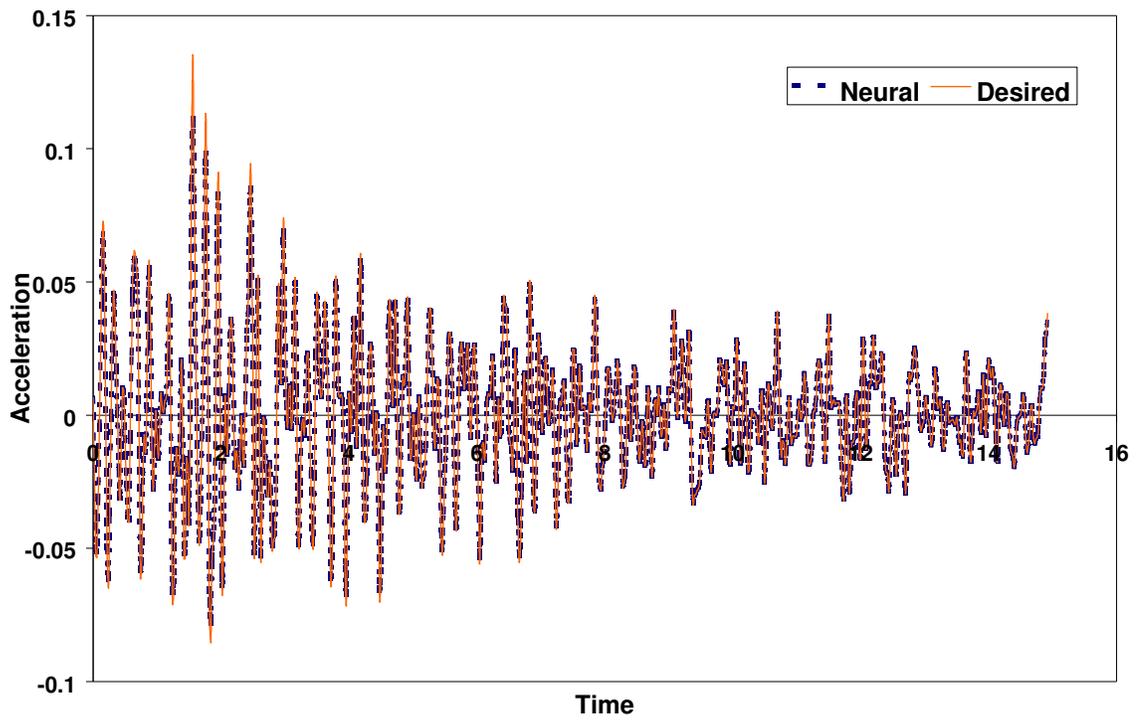

**Fig. 3(a). 80% Response comparison Between Neural and Desired of Chamoli Earthquake at Barkot (NE) for ω=0.01**

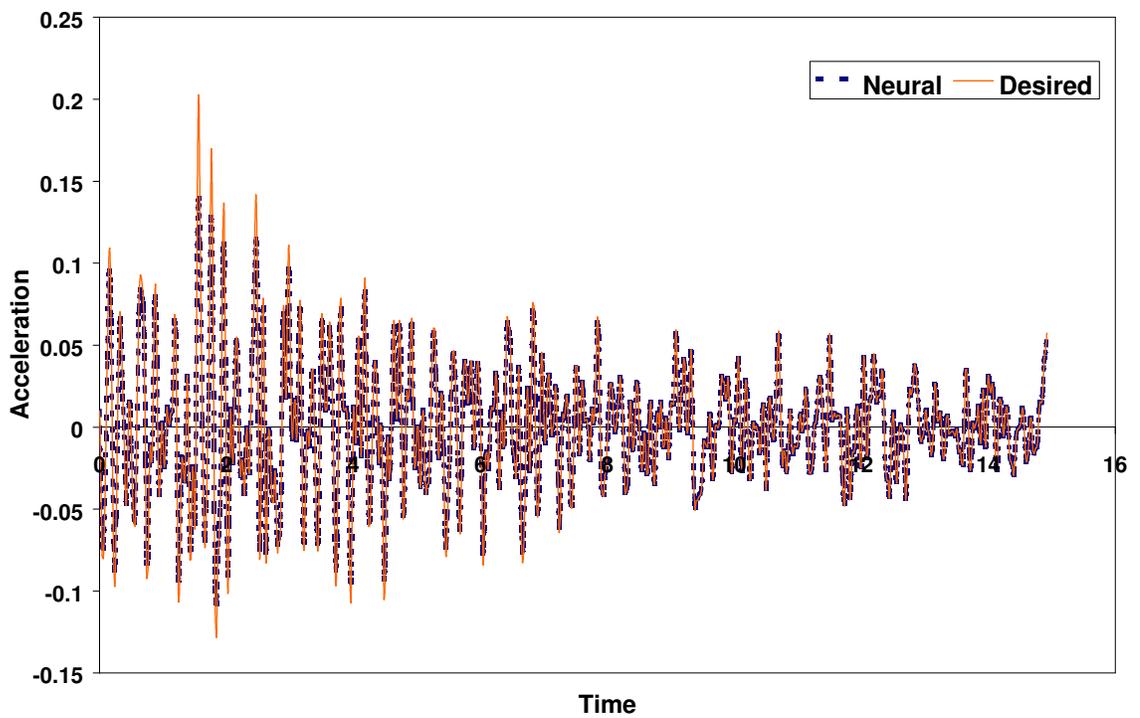

**Fig. 3(b). 120% Response comparison Between Neural and Desired of Chamoli Earthquake at Barkot (NE) for ω=0.5**

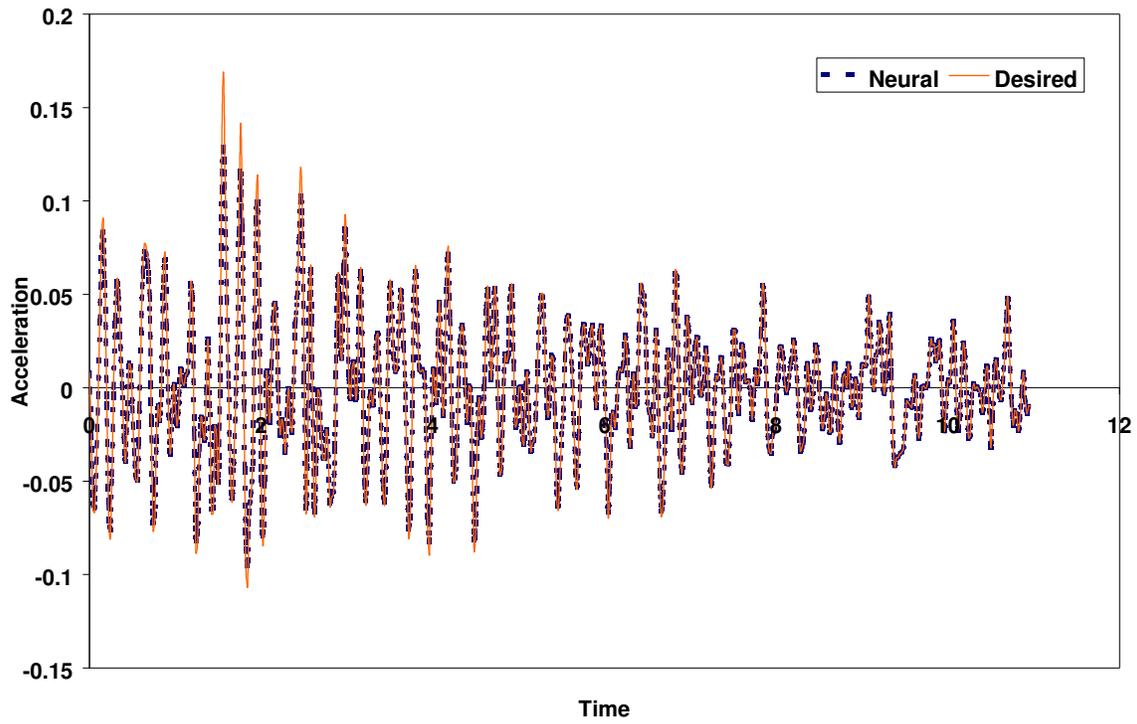

**Fig. 4(a). Response Comparison Between Neural and Desired (550 points) of Chamoli Earthquake at Barkot (NE) for ω=0.01**

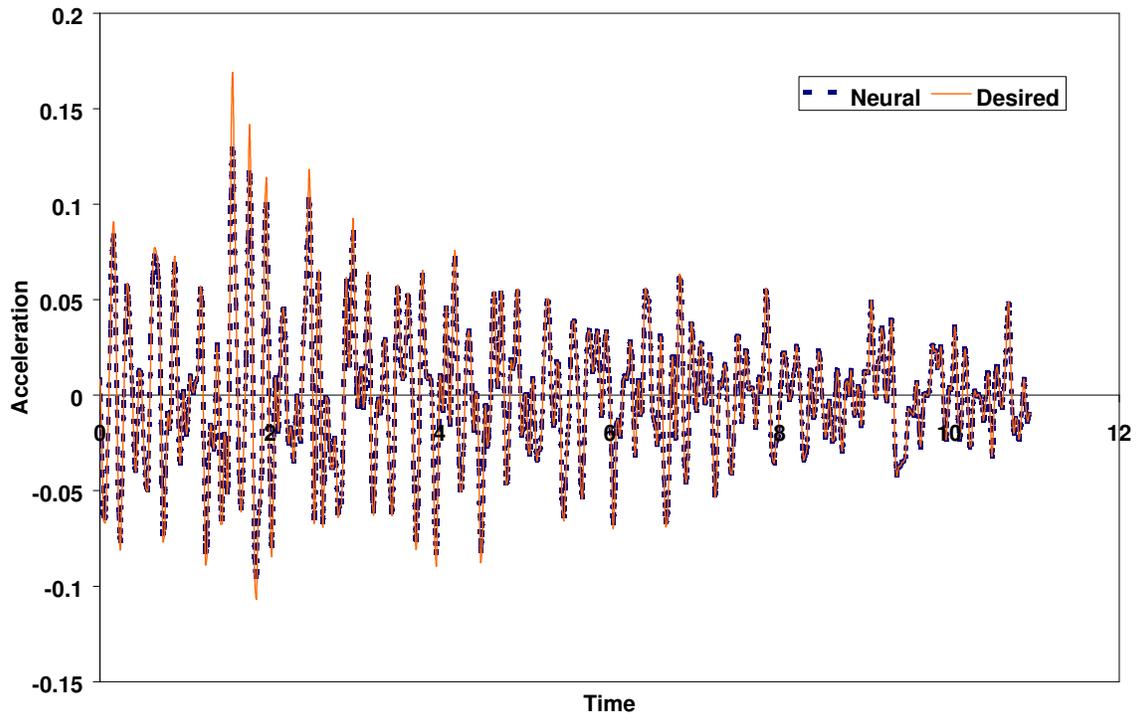

**Fig. 4(b). Response Comparison Between Neural and Desired (550 points) of Chamoli Earthquake at Barkot (NE) for ω=0.5**

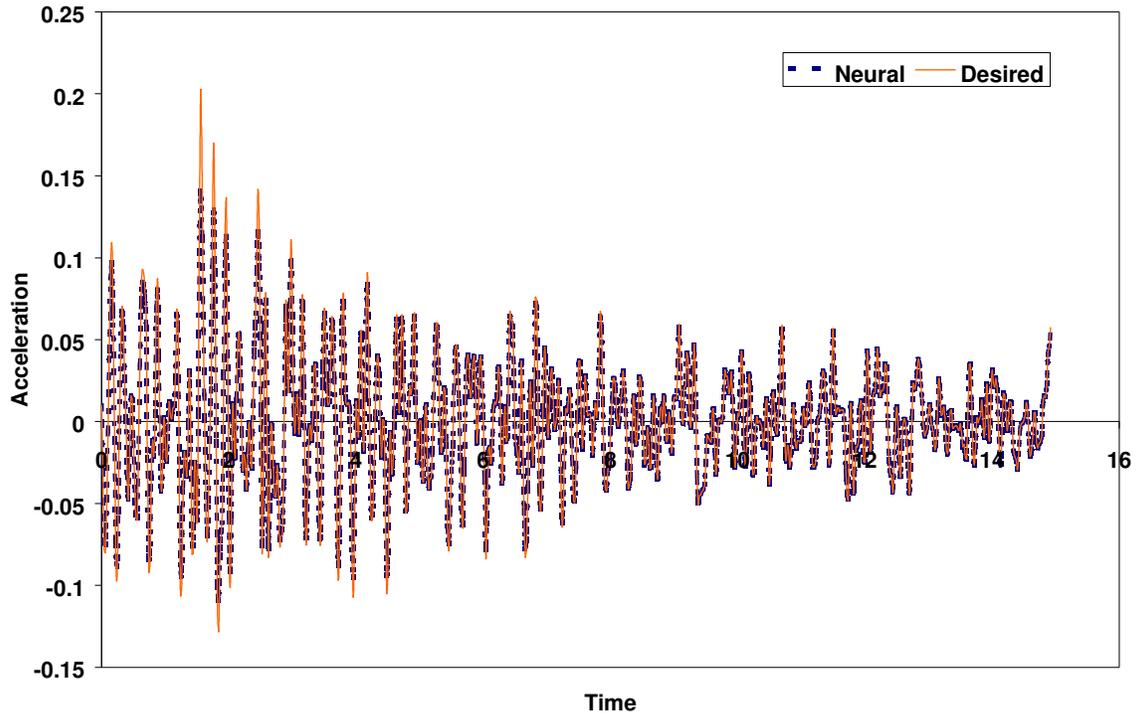

**Fig. 5(a). 120% Response Comparison Between Neural and Desired of Chamoli Earthquake at Barkot (NE) (748 points) ω=0.01 (After training from 550 points)**

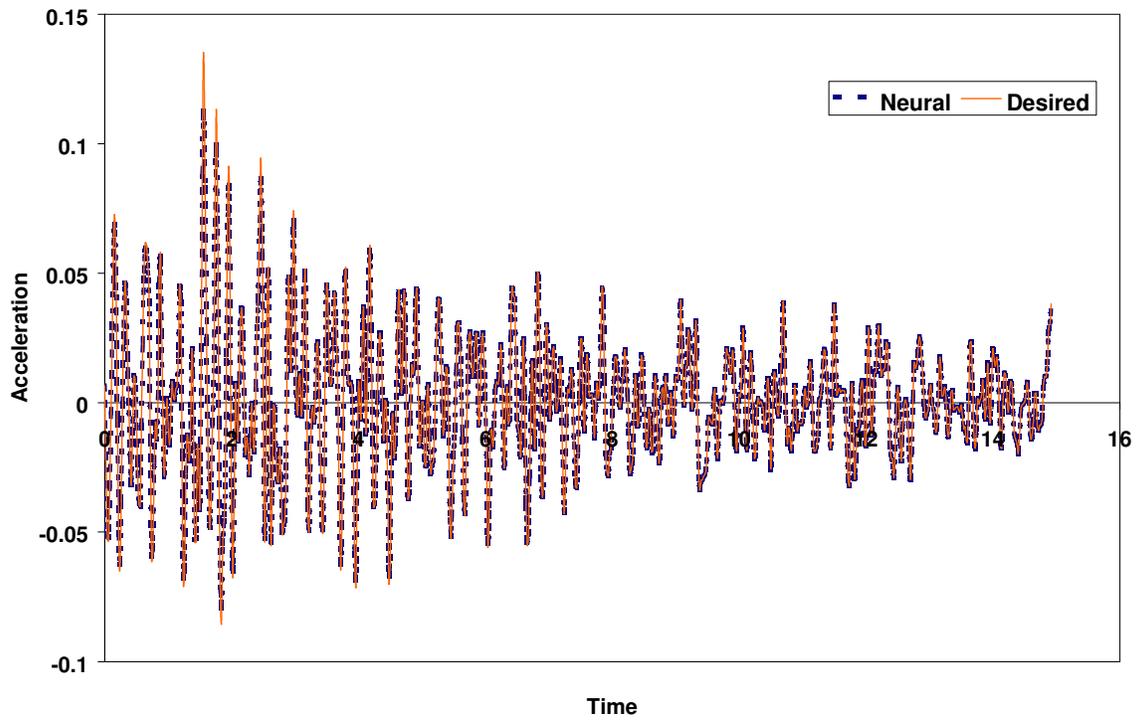

**Fig. 5(b). 80% Response Comparison Between Neural and Desired of Chamoli Earthquake at Barkot (NE) (748 points) ω=0.5 (After training from 550 points)**

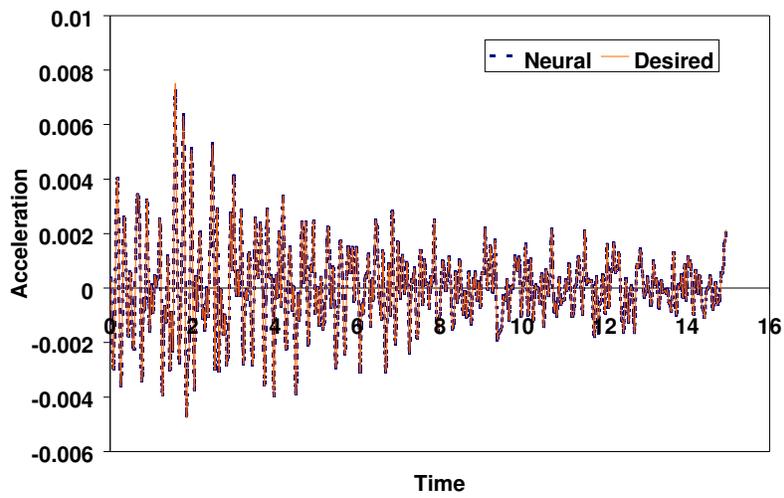

**Fig.6(a). 100% Response Comparison Between Neural and Desired of Chamoli Earthquake at Barkot (NE) with Damping**

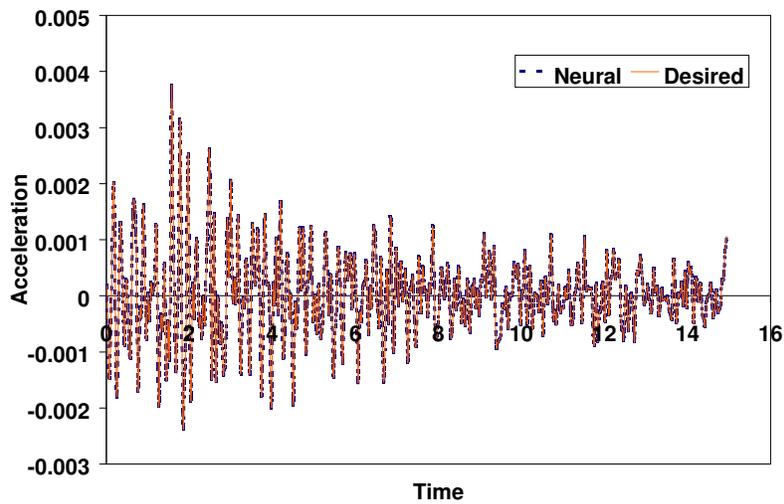

**Fig.6(b). 50% Response Comparison Between Neural and Desired of Chamoli Earthquake at Barkot (NE) with Damping**

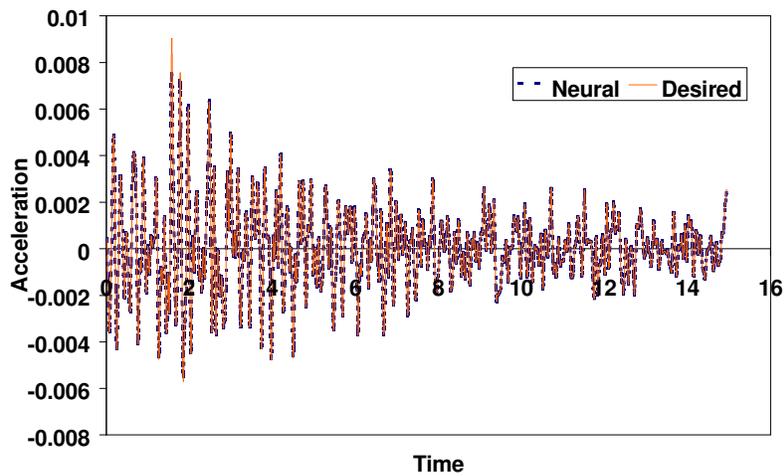

**Fig.6(c). 120% Response Comparison Between Neural and Desired of Chamoli Earthquake at Barkot (NE) with Damping**

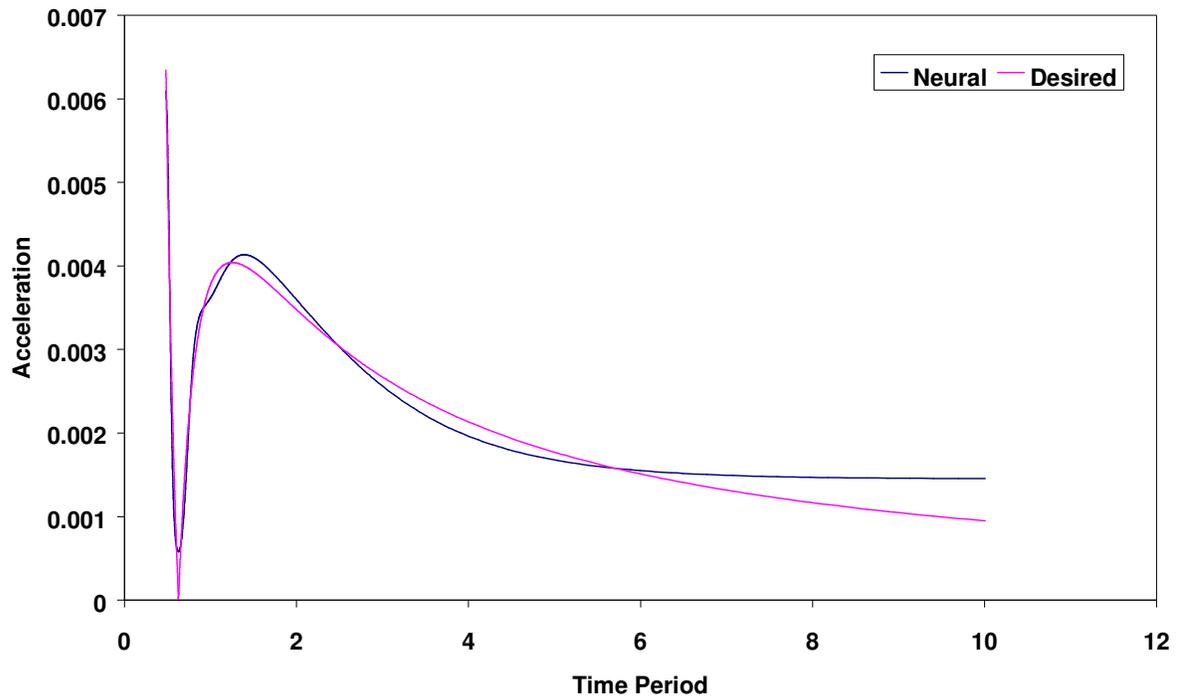

**Fig. 7(a). Comparison Between Neural and Desired for time period and the corresponding maximum response of Uttarkashi earthquake at Barkot in NW direction ( 620 points )**

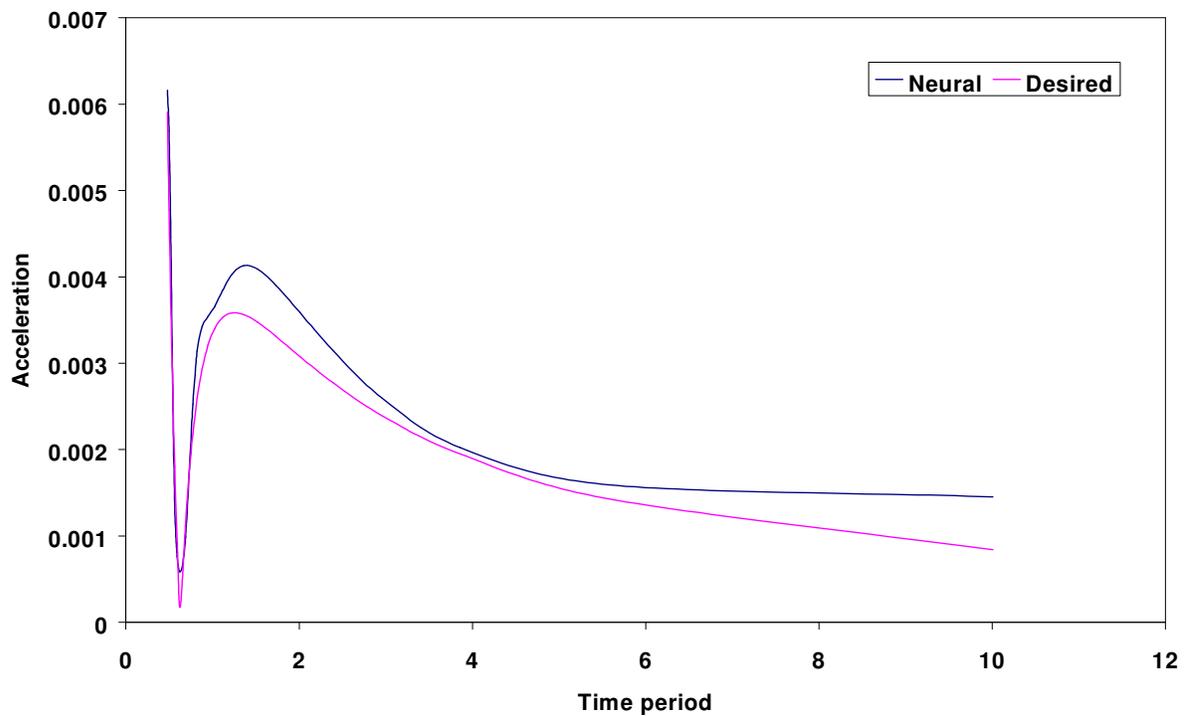

**Fig. 7(b). Comparison Between Neural and Desired for time period and the corresponding maximum response of Uttarkashi earthquake at Tehri in NW direction ( From weights of 620 points )**